\documentclass[conference]{IEEEtran}
\IEEEoverridecommandlockouts
\usepackage{cite}
\usepackage{amsmath,amssymb,amsfonts}
\usepackage{algorithmic}
\usepackage{algorithm}
\usepackage{graphicx}
\usepackage{textcomp}
\usepackage{xcolor}
\usepackage{caption}
\usepackage{subcaption}
\usepackage{dsfont}
\usepackage{tabularx}
\usepackage{multirow}
\usepackage{hyperref}
\usepackage{bbm}
\usepackage{array, makecell}
\usepackage{amsbsy}
\usepackage[compact]{titlesec}
\titlespacing{\section}{0pt}{*1}{*1}

\newcolumntype{C}{>{\centering\arraybackslash}X}
\def\BibTeX{{\rm B\kern-.05em{\sc i\kern-.025em b}\kern-.08em
    T\kern-.1667em\lower.7ex\hbox{E}\kern-.125emX}}

\newcommand{\eg}{\textit{e}.\textit{g}.}
\newcommand{\ie}{\textit{i}.\textit{e}.}
\definecolor{blue(ncs)}{rgb}{0.0, 0.53, 0.74}
\definecolor{applegreen}{rgb}{0.55, 0.71, 0.0}
\begin{document}

\title{MOGNET: A Mux-residual quantized Network leveraging Online-Generated weights \\

}
\author{\IEEEauthorblockN{Van Thien Nguyen, William Guicquero and Gilles Sicard}
CEA-LETI, F-38000, Grenoble, France \\
\{vanthien.nguyen, william.guicquero, gilles.sicard\}@cea.fr}

\makeatletter
\def\ps@IEEEtitlepagestyle{
  \def\@oddfoot{\mycopyrightnotice}
  \def\@evenfoot{}
}
\def\mycopyrightnotice{
  {\footnotesize
  \begin{minipage}{\textwidth}
  \centering
  Copyright~\copyright~2022 IEEE. Personal use of this material is permitted. However, permission to use this material \\ 
  for any other purposes must be obtained from the IEEE by sending an email to pubs-permissions@ieee.org. DOI: \href{https://doi.org/10.1109/AICAS54282.2022.9869933}{10.1109/AICAS54282.2022.9869933}
  \end{minipage}
  }
}

\maketitle

\begin{abstract}
This paper presents a compact model architecture called MOGNET, compatible with a resource-limited hardware. MOGNET uses a streamlined Convolutional factorization block based on a combination of 2 point-wise ($1\times1$) convolutions with a group-wise convolution in-between. To further limit the overall model size and reduce the on-chip required memory, the second point-wise convolution's parameters are on-line generated by a Cellular Automaton structure. In addition, MOGNET enables the use of low-precision weights and activations, by taking advantage of a Multiplexer mechanism with a proper Bitshift rescaling for integrating residual paths without increasing the hardware-related complexity. To efficiently train this model we also introduce a novel weight ternarization method favoring the balance between quantized levels. Experimental results show that given tiny memory budget (sub-2Mb), MOGNET can achieve higher accuracy with a clear gap up to 1$\%$ at a similar or even lower model size compared to recent state-of-the-art methods.
\end{abstract}


\section{Introduction}
The successful use of Convolutional Neural Networks (CNNs) in image recognition tasks has been recently accompanied by a considerable increase in model architectures complexity, expanding the number of parameters as well as the computational costs. Unfortunately, this limits the deployment of such network models in embedded systems with limited hardware resources. Therefore, designing lightweight models --regarding memory and computational capabilities-- is a challenge to enable accurate inference tasks at the edge. Recent efforts towards alleviating this algorithmic overhead involve several techniques such as efficient model design \cite{DWSConv}, network quantization \cite{QNN} and layer inter-connection pruning \cite{Magpruning}.    

Our goal here is to reduce the overall hardware needs required to run a model implemented in resource-constrained devices (\eg, for ASIC design) while still ensuring an acceptable accuracy. Unlike several works focusing on large models to achieve extremely high compression rates \cite{DBLP:journals/corr/HanMD15}, \cite{DBLP:conf/iclr/OktayBSS20}, we first propose a hardware-compliant model architecture to which we further apply efficient quantization methods.        

In this paper, we present the compact MOGNET model architecture which combines:
\begin{itemize}
\item quantized residual modules with a Multiplexer-based skip mechanism and,
\item a custom factorization of convolution layers that uses on-line generated weights.
\end{itemize}
Indeed, a Cellular Automaton (CA) is used to automatically generate the weights of a pointwise convolution in each factorized-CNN block, thus reducing parameter-related storage requirements. Moreover, we introduce a novel training framework to obtain the ternary weights in our model which favors the balance between $3$ discrete levels. 

\section{Related works}

\textbf{Residual connections} \cite{ResNet} have become important elements of modern CNN architectures, which aim at increasing model expressivity, favoring feature reuse, and alleviating the gradient vanishing in deep CNNs. \cite{ResAtt} then incorporates an attention mechanism into the residual learning. SENet \cite{SeNet} proposes a channel attention involving feature aggregation and recabliration stages. MOGNET also employs these aforementioned concepts, however, it mainly focuses on the possible hardware mapping of the model. While previous works perform the residual connections using full-precision, we propose a quantized Multiplexer (MUX) layer with BitShift rescaling allows integrating both an addition connection and a channel-wise attention-like mechanism with a very limited data precision and thus restrained hardware-related costs.    

\textbf{Quantization} reduces the precision of weights and activations for low-bitwidth computations. Advanced methods integrate the quantization during training to jointly optimize the quantizer with the quantized weights, under the minimization of the quantization error \cite{Li2016TernaryWN}, \cite{Zhao2020LinearSQ} or the output task loss \cite{EsserMBAM20}. In MOGNET, for the sake of versatility, quantizers are regularized to provide balanced quantized outputs.


\textbf{Efficient architecture design} involves the definition of alternative architectures to the hardware-expensive canonical network models. Depthwise Separable Convolution (DSConv \cite{Sifre14}) has become a building block in different CNN architectures \cite{MobileNetsEC}, \cite{DWSConv} which performs a depthwise convolution followed by a pointwise ($1\times1$) convolution. \cite{BSConv} revisits DSConv by putting the pointwise (with orthonormal regularization) before the depthwise convolution to cap the redundancy. Grouped convolution is also introduced \cite{ShuffleNet} on the continuum between regular convolution and DSConv. ResNext \cite{AggregatedRT} presents a building block including 2 pointwise layers and a grouped convolution inserted in-between. Our work proposes a factorization similar to the building block of ResNext, without activation inserted in-between layers.

\textbf{Cellular Automata} \cite{CA} enable the generation of random states given an initialization and update rule function. Hence, this repetitive structure is commonly used for generating on-the-fly sets of random vectors \cite{Liu2017ChaoticCA}, or computing in random representations \cite{Ylmaz2014ReservoirCU}. In MOGNET, we leverage CA to generate part of weight parameters of our model, consequently reducing the overall model memory-related footprint. 

\begin{figure}[htbp]
	\centerline{\includegraphics[scale=0.25]{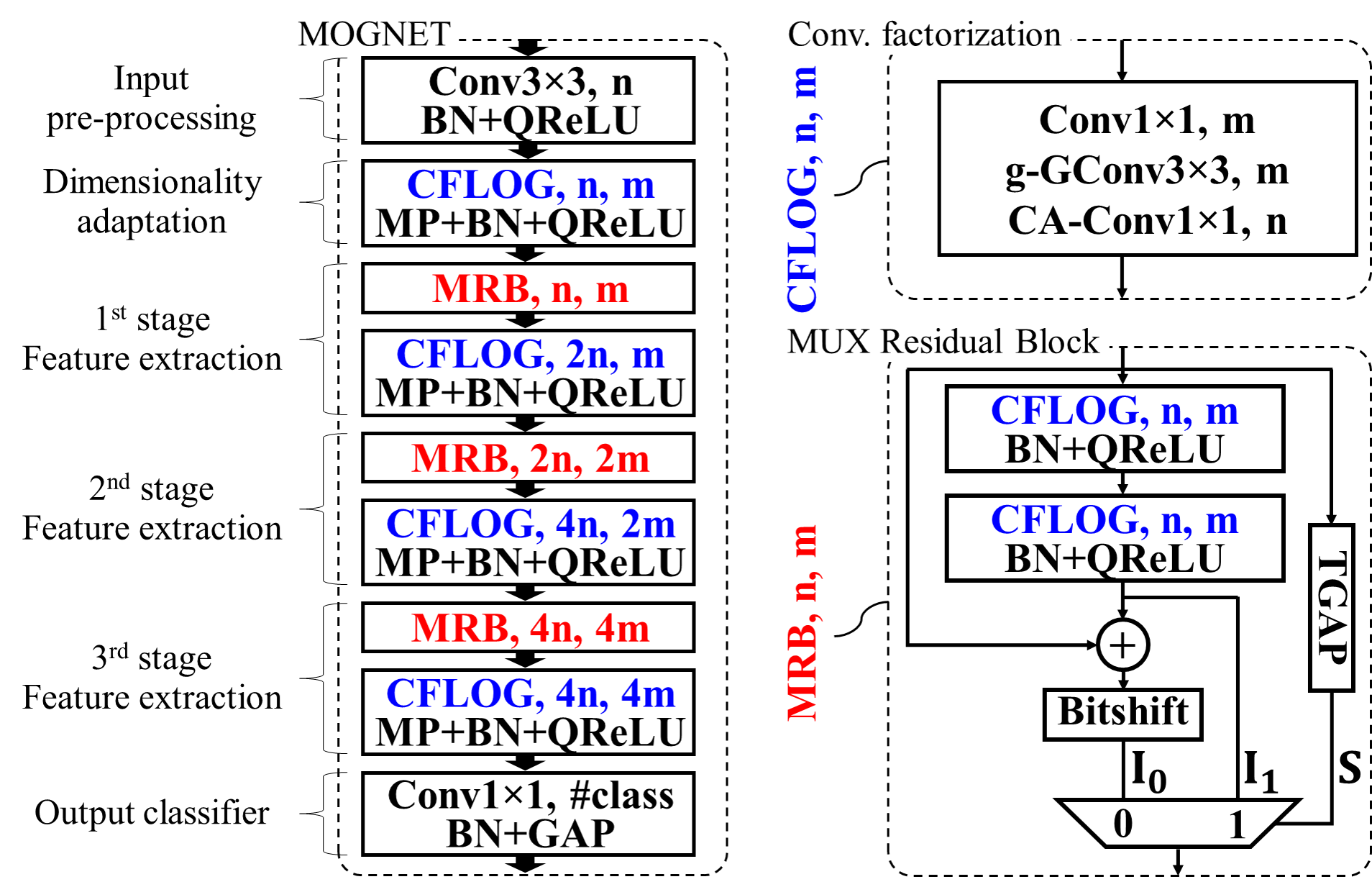}}
	\caption{Top-level architecture description of MOGNET with Convolutional Factorization Leveraging On-line Generated weights (\textcolor{blue}{CFLOG}) and MUX Residual Block (\textcolor{red}{MRB}). The final 1$\times$1 convolution is followed by Batch Normalization (BN) prior to a Global Average Pooling (GAP). Here $n, m$ are the parameters controlling the number of output feature maps and the latent dimension in \textcolor{blue}{CFLOG}, MP stands for 2$\times$2 Max Pooling and g-GConv is Grouped Convolution with g groups.} 
	\label{MOGNET}
\end{figure}

\section{MOGNET}
Figure~\ref{MOGNET} describe the MOGNET architecture that uses integer-only multiplication-accumulations (MACs) and hardware-compliant operations such as 1-bit Bitshifts and 2-input multiplexers. The following description yet presents MOGNET from its algorithmic view point, \ie \, with computations done in a real-valued domain but with relevant hardware-equivalent specializations. In this section, we first focus on our custom MUX Residual Block (MRB), then on our convolution layer factorization (CFLOG).

\subsection{MUX residual Block (MRB)}
We denote $k$ as the quantization bitwidth of the activations throughout the network. Indeed, a $k$-bit Quantized Rectified Linear Unit (QReLU) is defined so that for any input, the outputs are in the set $\{0, \frac{1}{2^k-1}, \frac{2}{2^k-1}, ... 1\}$:
\begin{equation}
\textrm{QReLU} (x; k) = \begin{cases}
 \frac{1}{2^k - 1} \lfloor (2^k - 1) x \rceil & \textrm{if } k > 1 ,\\
\mathds{1}_{\{x > 0\}} & \textrm{if } k = 1.
\end{cases}
\label{quant_relu}
\end{equation}

For the backward pass, we compute the gradient using the Straight-Through-Estimator strategy (STE, \cite{bengio_estimating_2013}) $\frac{\partial \textrm{QReLU}}{\partial \boldsymbol{x}} = \mathds{1}_{\{|\boldsymbol{x}|\leq 1\}}$. As the Addition $y = x_1 + x_2$ of two unsigned $k$-bit activations $x_1$, $x_2$ will increase the dynamic range by 1-bit, we make use of the following Bitshift software description inside the MRB, to keep $y$ always at $k$-bit: 

\begin{equation}
\textrm{Bitshift} (y; k) = \begin{cases}
 \frac{1}{2^k - 1} \lfloor \frac{(2^k - 1) y}{2} \rfloor & \textrm{if } k > 1,\\
 \lceil \frac{y}{2} \rceil & \textrm{if } k = 1.
\end{cases}
\label{bitshift}
\end{equation}

Due to the specific use of this function for the add-type connection, we adopt a completely-passed-through gradient $\frac{\partial \textrm{Bitshift}}{\partial \boldsymbol{y}} = 1$. For $k>1$, this rescaling can be implemented by a 1-bit bitshift, while in the specific case of $k=1$, the combination of the addition and the bitshift can be replaced by an appropriate single \textbf{OR} gate. Let us denote $\textbf{I}_0, \textbf{I}_1 \in \mathbb{R}^{h\times w \times n}$ as the output of the Bitshift operation and the second QReLU where $h, w, n$ are the height, width and number of channels; $\textbf{S} \in \{0, 1\}^{1\times1\times n}$ as the binary control signal. The MRB core element is \textbf{MUX} which can be mathematically described as:

\begin{equation}
\textrm{\textbf{MUX}} (\textrm{\textbf{I}}_0, \textrm{\textbf{I}}_1; \textrm{\textbf{S}}) = \textrm{\textbf{I}}_1 \odot \textrm{\textbf{S}} + \textrm{\textbf{I}}_0 \odot (\textbf{1} - \textrm{\textbf{S}})   
\label{mux}
\end{equation}
where $\odot$ is a channel-wise multiplication. 
The control signal $\textbf{S}$ embeds a parameter-free channel attention which consists in a Thresholded Global Average Pooling (TGAP). TGAP simply corresponds to a channel-wise Global Average Pooling (GAP) followed by a binarization $T(x) = \mathds{1}_{\{x > 0.5m\}}$, where $m$ is set to the maximum of the GAP's outputs in the full-precision representation and to $1$ in the quantized model which is the maximum possible value of quantized activations. For the hardware deployment, this last operation can be implemented via an integer accumulation followed by an integer-to-integer comparison. This way, the input of each MRB will automatically control the operation of the Multiplexer module in a channel-wise manner. Concretely, MRB will perform the Additional connection for each input feature map that is dominated by small values. Otherwise, the MRB will simply keep the straightforward output of the second QReLU. One interesting aspect of this MUX-skip connection is that it favors the balance between the number of large-valued data with respect to the number of small-valued data throughout the networks, this without any other regularization strategy.

\subsection{Convolution factorization leveraging CA-generated weights}
\begin{figure}[htbp]
	\centerline{\includegraphics[scale=0.25]{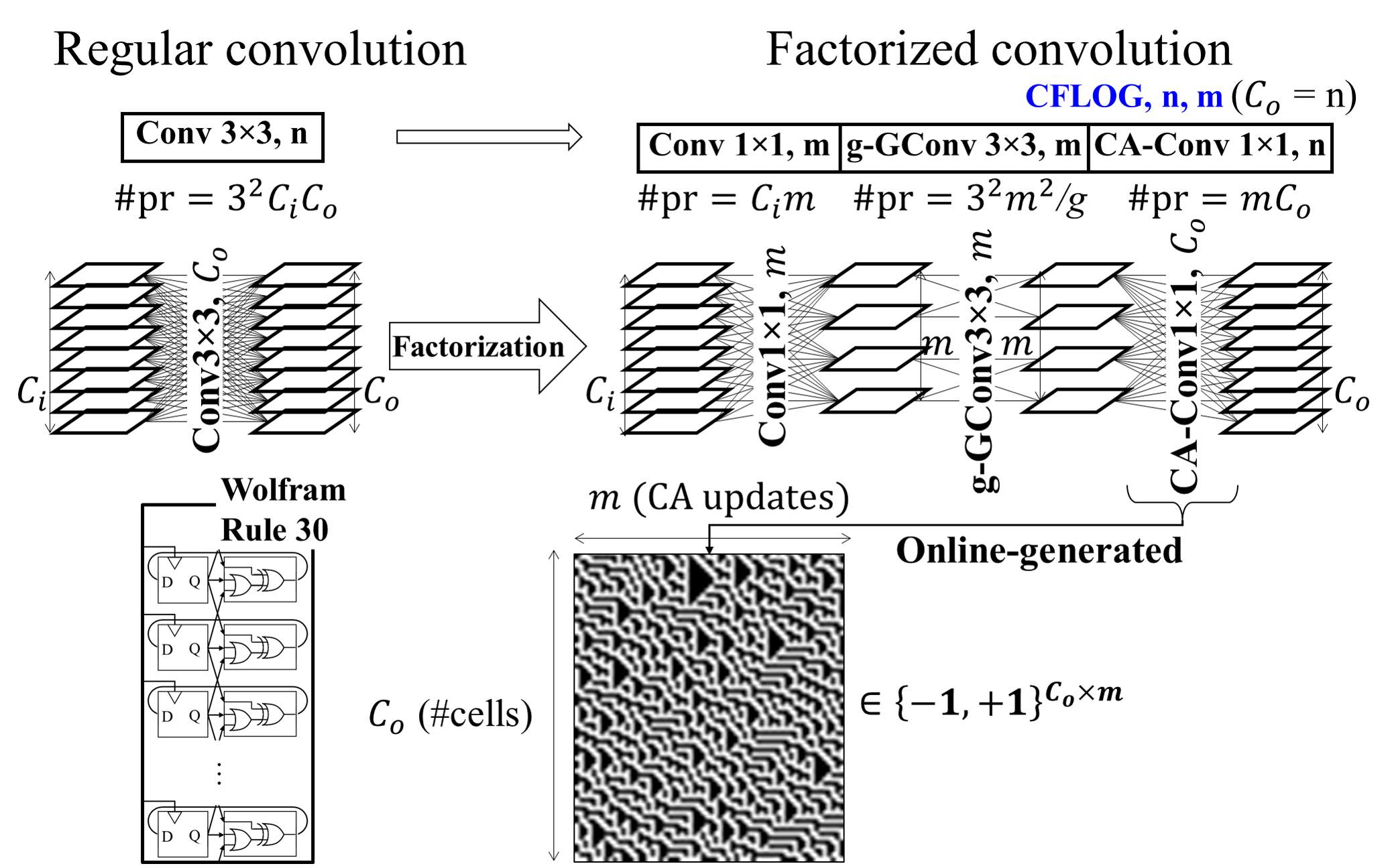}}
	\caption{CFLOG description with CA-generated weights.} 
	\label{PGCA}
\end{figure}

To further reduce the on-chip memory and the computational complexity of the model, we replace all regular convolutions (except the first and the last layers, cf. Fig.~\ref{MOGNET}) by a light-weight factorization consisting of 2 pointwise layers and a grouped convolution (Fig.~\ref{PGCA}), namely CFLOG. Unlike the building block in ResNext \cite{AggregatedRT}, we do not use any nonlinearity (\eg \, normalization, activation) between these layers. Moreover, to further reduce the model size, the last pointwise convolution's weights are fixed during training and generated in real-time by a CA given a certain seed. The first pointwise convolution embeds the input feature into low-dimension $m < C_{i}$, the number of input channels. The grouped conv layer performs $g$ groups of convolutions and also outputs $m$ feature maps. Finally, these feature maps are sequentially projected back to a high-dimensional space of $C_{o}$ channels ($C_o = n$ in \textcolor{blue}{CFLOG, n, m}) thanks to a CA-generated kernel. As depicted in Fig.~\ref{PGCA}, this kernel is formed by concatenating all states obtained when evolving a $C_o$-cell CA during $m$ update states. In this work, we consider Wolfram's rule 30 for the local evolution function between states. We choose $m=\frac{C_{i}}{2}$ that gives the following compression rate (CR) between the number of trainable parameters ($\#pr$) of CFLOG and that of the regular convolution:
\begin{equation}
\textrm{CR} = \frac{C_{i}m + \frac{3^2m^2}{g}}{3^2C_{i}C_{o}} = \frac{C_{i}}{C_{o}} \left( \frac{1}{18} + \frac{1}{4g}\right)
\label{cr}
\end{equation}

\subsection{Balanced Ternary Quantization (BTQ)}
\begin{figure}[h]
	\centerline{\includegraphics[scale=0.2]{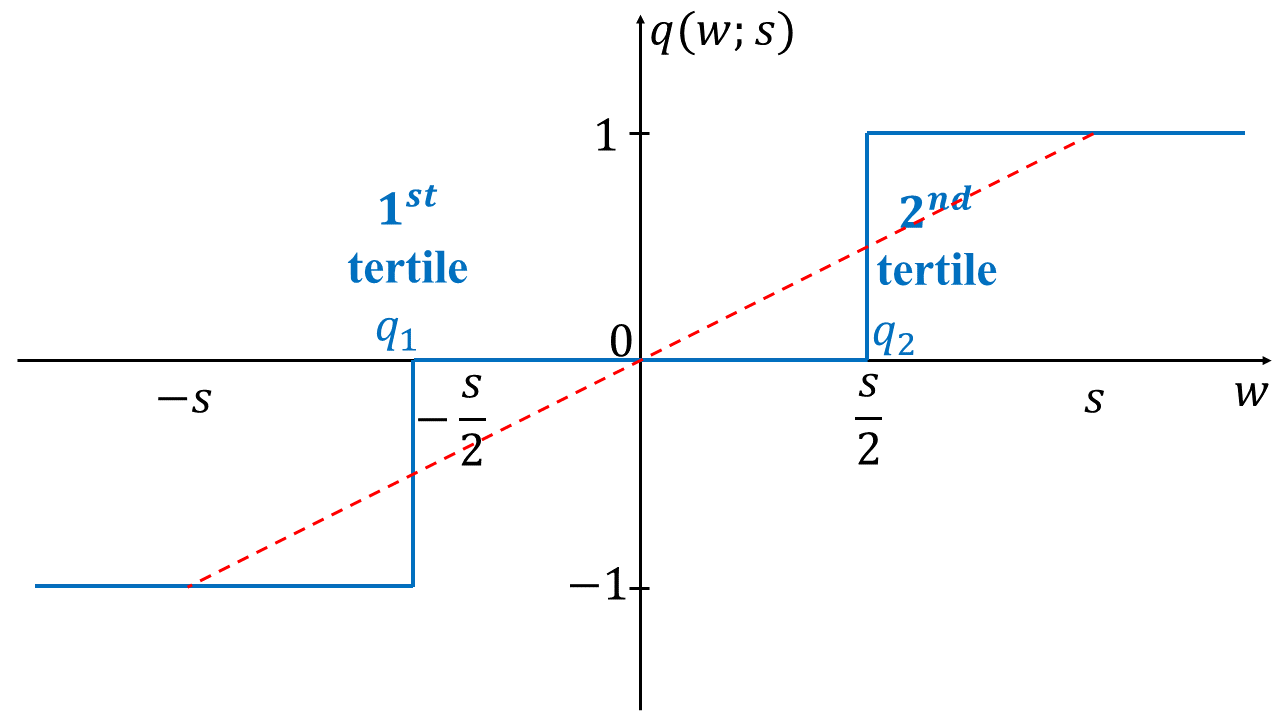}}
    \caption{Balanced ternary quantization with histogram bin equalization when 2 tertiles ($q_1, q_2$) are symmetrical and coincide with the quantization thresholds ($-\frac{s}{2}, \frac{s}{2}$).} 
	\label{heq}
\end{figure}

In order to drastically compress the model, we binarize \cite{BNN} all learnable pointwise and ternarize all other layers' weights. To obtain the ternarized parameters, we introduce a novel quantization-aware training scheme which favors the balance between $3$ discrete levels. The ternary mapping $q: \mathbb{R} \rightarrow \{-1, 0, +1\}$ is applied to the real-valued proxy weights $\boldsymbol{w}$,
\begin{equation}
q(w; s) =  \textrm{Clip}\left( \left \lfloor\frac{w}{s} \right \rceil, -1, 1 \right),
\label{hepq}
\end{equation}
where $s$ is the step size parameter. We still adopt the STE gradient $\frac{\partial q}{\partial \boldsymbol{w}} = \mathds{1}_{\{|\boldsymbol{w}|\leq 1\}}$. 
In Fig.~\ref{heq} we denote $q_1$ and $q_2$ as the \textbf{tertiles} of the proxy weight's histogram. Observing that the proxy weights distribution may change during the training, but the median value usually stays around zero, we assume that $q_1$ and $q_2$ are symmetrically distributed, \ie \, $q_{1} \approx - q_2$ with $q_2 > 0$. Therefore, to equi-distribute the quantized weights, we can automatically update the step size $s$ at the beginning of each epoch based on $q_1$ and $q_2$. Concretely, we expect that the sum of the absolute value of the tertiles ($q_1, q_2$) is approximately equal to that of the thresholds ($-\frac{s}{2}, \frac{s}{2}$), therefore we have:
\begin{equation}
s = |q_1| + |q_2|.
\label{het}
\end{equation}
The algorithm for training BTQ is detailed in Algorithm~\ref{algo}, in which the step size update is described in lines $2-5$.      
\begin{algorithm}
 \caption{Training MOGNET with BTQ}
 \begin{algorithmic}[1]
 \label{algo}
 \renewcommand{\algorithmicrequire}{\textbf{Input:}}
 \renewcommand{\algorithmicensure}{\textbf{Output:}}
\REQUIRE Initial proxy weights $\{\textbf{W}_l\}_{l=1}^L$ and training dataset
 \ENSURE  Optimized $\{\textbf{W}_l\}_{l=1}^L$,  $\{s_l\}_{l=1}^L$ \\
 // $B$, $T$, $L$: \#batches, \#epochs, \#layers using BTQ\\
 // $\textbf{W}_l$: full-precision proxy weights of the $l^{th}$ layer,\\
 // $s_l$: quantization step size used at the $l^{th}$ layer, \\
  \FOR {$t = 1$ to $T$} 
  \FOR {$l = 1$ to $L$}
  \STATE Find $2$ tertiles $q_1$ and $q_2$ of layer $l$ 
  \STATE Compute and update $s_l$ (Eq.~\ref{het})
  \ENDFOR
  \FOR{$b = 1$ to $B$}
  \STATE Forward pass using $\{q(\textbf{W}_l; s_l)\}_{l=1}^L$ (Eq.~\ref{hepq})
  \STATE Backward pass and update $\{\textbf{W}_l\}_{l=1}^L$ 
  \ENDFOR
  \ENDFOR
 \end{algorithmic} 
 \end{algorithm}
 
\section{Experiments}
\subsection{Experimental settings}
We implemented all the proposed elements using the TensorFlow \cite{TensorFlow} library and CellPylib \cite{Antunes2021} package. The quantized models are first initialized from their full-precision counterparts being trained on CIFAR-10 and CIFAR-100 \cite{krizhevsky_learning_nodate} datasets from scratch, where ReLU and Linear activations replace our QReLU and BitShift. Then, we train the quantized models through a 2-stage procedure: first train quantized weights with full-precision activations and second, fine-tune the quantized weights with all quantized activations. We apply the simple data augmentation scheme for training: random crop from all-sided 4-pixel padded images combined with random horizontal flips. Table~\ref{training_settings} details the training and optimization setting used to derive our experimental results. 

\begin{table}[h]
\caption{Training and optimization settings}
\centering
\begin{tabular}{| c | c | c |}
\hline
Dataset & CIFAR-10 & CIFAR-100 \\
\hline
Optimizer & \makecell{Adam \cite{AdamOp} \\ ($\beta_1 = 0.9, \beta_2=0.999$)} & \makecell{Adam \\ ($\beta_1 = 0.9, \beta_2=0.999$)} \\
\hline
Initial learning rate (LR) & $10^{-3}$ & $10^{-3}$ \\
\hline
Batch size &  50  & 50 \\
\hline
First stage epoch &    180   &  250        \\
\hline
\makecell{First stage \\LR schedule} & \makecell{Exponentially decay\\ after 120-th epoch}     &  \makecell{Exponentially decay\\ after 150-th epoch}      \\
\hline
Second stage epoch &    150   &  250        \\
\hline
\makecell{Second stage \\LR schedule} &  \makecell{Exponentially decay \\ after 80-th epoch}      &   \makecell{Exponentially decay \\ after 150-th epoch}      \\
\hline
Rate of LR decay & 0.9 & 0.9 \\
\hline
\end{tabular}
\label{training_settings}
\end{table}

\subsection{Experimental results}
\begin{figure}[h]
     \centering
         \begin{subfigure}[b]{0.24\textwidth}
        \includegraphics[width=\textwidth]{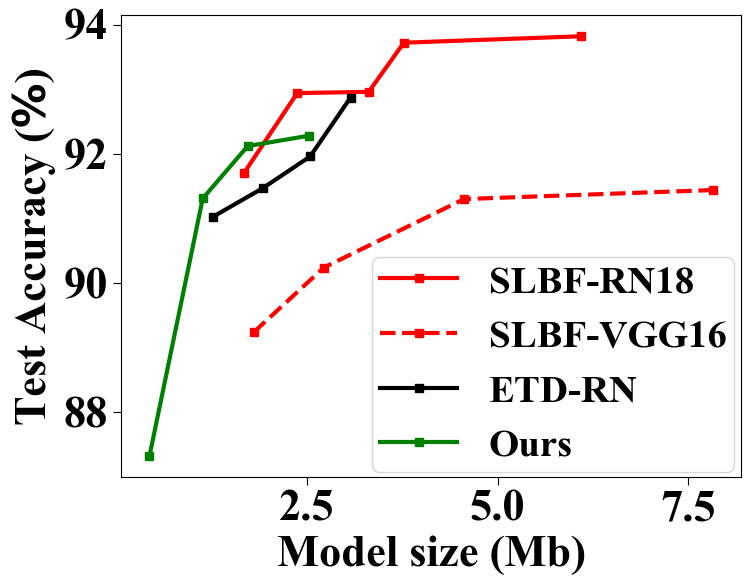}
        \caption{CIFAR-10.}
        \label{c10}
    \end{subfigure}
    \hspace{-2mm}
    \begin{subfigure}[b]{0.24\textwidth}
        \includegraphics[width=\textwidth]{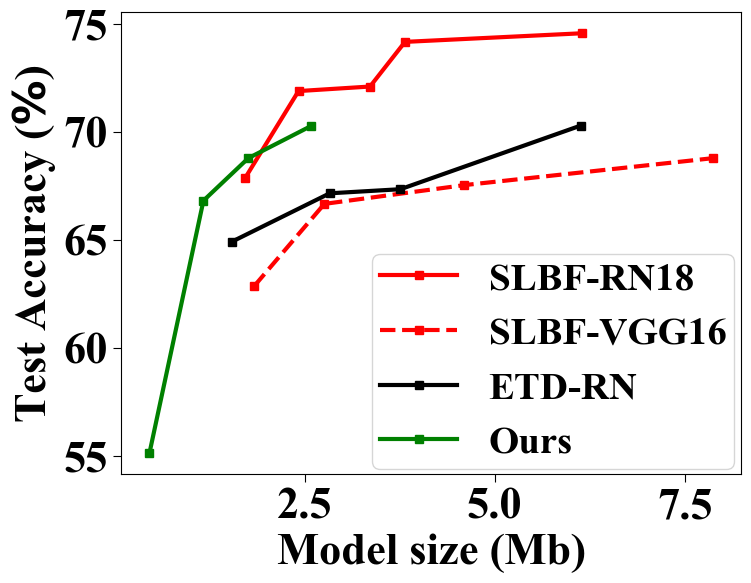}
        \caption{CIFAR-100.}
        \label{c100}
    \end{subfigure}
     \caption{Test accuracy of different compression method-model couplings. Our models are with 3-b activations.}
     \label{acc_curves}
\end{figure}

We evaluate the performance of our models in comparison with recent state-of-the-art model compression techniques: Stacking Low-dimensional Binary Filters (SLBF \cite{Lan2021CompressingDC}) on ResNet (RN)-18 and VGG-16 \cite{VGG}; Efficient Tensor Decomposition (ETD\cite{ETD}) on RN-20 and RN-32. Table~\ref{sota_table} reports the model size, the activation precision as well as the accuracy of different methods and models. It demonstrates that for $n=128$ and $g=4$, MOGNET achieves the highest accuracy level on CIFAR-10, while having lower model size and 3-bit only activations. Moreover, at the same configuration, MOGNET outperforms other methods on CIFAR-100 with a clear gap of nearly $1\%$ ($67.89 \rightarrow 68.80 \%$) at the similar weight-related memory. We can mention the impact of the hyperparameter $k$ on the model performance, with a significant degradation when decreasing the activation precision to 1-bit or 2-bit. Figure~\ref{acc_curves} additionally reports accuracy versus model size curves of various considered models/hyperparameters. On both two datasets, MOGNET stays in the optimal top-left zone implying low on-chip memory requirements with high accuracy. However, when increasing the model size ($n > 128$), MOGNET curves fall under that of SLBF-RN18. This last result means that MOGNET, with its limited depth, is more relevant to target extremely low-sized ($<2$Mb) models.

\begin{table}[h]
\caption{Comparison of different network compression methods on CIFAR-10 and CIFAR-100.}
\centering
\begin{tabular}{| c | c | c | c | c | c |}
\hline
\multirow{2}{*}{\makecell{\textbf{Method} \\ Model}} & \multirow{2}{*}{\makecell{Activation \\ Bitwidth \\ ($k$)}} &\multicolumn{2}{c|}{CIFAR-10} & \multicolumn{2}{c|}{CIFAR-100} \\
    \cline{3-6}
     &   &  \makecell{Model size \\ (Mb)}   & Acc ($\%$) &   \makecell{Model size \\ (Mb)} & Acc ($\%$) \\
\hline
\textbf{\textbf{SLBF}\cite{Lan2021CompressingDC}} &     &       &        &    &  \\
RN-18 &  32 & 1.67 & 91.70 & 1.72 & 67.89 \\
VGG-16 & 32 & 1.84 & 89.24 & 1.89 & 62.88 \\
\hline
\textbf{\textbf{ETD}\cite{ETD}} &     &       &        &    &  \\
RN-20 &  32 & 1.94 & 91.47 & 3.80 & 67.36 \\
RN-32 & 32 & 2.54 & 91.96 & 2.83 & 67.17 \\
\hline
 \multirow{3}{*}{\makecell{\textbf{Ours} \\ ($n$=$128$, $g$=$8$)}} &  1 & \multirow{3}{*}{1.13} & 87.60  & \multirow{3}{*}{1.22}  & 59.30 \\
  &  2 &  & 90.81  &   & 65.88 \\
   & 3 &  & 91.31  &   & 66.83 \\
\hline
 \multirow{3}{*}{\makecell{\textbf{Ours} \\ ($n$=$128$, $g$=$4$)}} &  1 & \multirow{3}{*}{\textbf{1.72}} & 88.99  & \multirow{3}{*}{\textbf{1.76}}  & 61.27 \\
  &  2 &  & 91.16  &   & 66.55 \\
   & \textbf{3} &  & \textbf{92.12}  &   & \textbf{68.80} \\
   
\hline
\end{tabular}
\label{sota_table}
\end{table}

\section{Conclusion}
This work introduces a novel hardware-compliant quantized model architecture called MOGNET, which integrates a custom Multiplexer mechanism and a lightweight convolution factorization that leverages Cellular Automaton-generated weights, in order to limit the hardware memory needs. We empirically show that our method can achieve better accuracy with lower model size than previous works, in particular for a tiny memory budget, while limiting the digital dynamic range using 3-bit quantized activations for integer-only MACs.  

\bibliographystyle{IEEEtran}

\end{document}